\documentclass{article}

     \usepackage[final]{neurips_2020}

\usepackage[utf8]{inputenc} % allow utf-8 input
\usepackage[T1]{fontenc}    % use 8-bit T1 fonts
\usepackage{hyperref}       % hyperlinks
\usepackage{url}            % simple URL typesetting
\usepackage{booktabs}       % professional-quality tables
\usepackage{amsfonts}       % blackboard math symbols
\usepackage{nicefrac}       % compact symbols for 1/2, etc.
\usepackage{microtype}      % microtypography
\usepackage{amsmath}
\usepackage{graphicx}
\usepackage{subfigure}
\usepackage{booktabs}
\usepackage{multirow}
\usepackage{color}
\usepackage{authblk}
\usepackage{hyperref}

\newcommand{\etal}{\textit{et al}. }
\newcommand{\ie}{\textit{i}.\textit{e}. }

\title{Unfolding the Alternating Optimization for Blind Super Resolution}

\author[1,2,3]{Zhengxiong Luo}
\author[1,2]{\thanks{Corresponding author} Yan Huang}
\author[2,3]{Shang Li}
\author[1,4,5]{Liang Wang}
\author[1,4]{Tieniu Tan}

\affil[1]{
	Center for Research on Intelligent Perception and Computing (CRIPAC) \authorcr
	National Laboratory of Pattern Recognition (NLPR)
}
\affil[2]{
	Institute of Automation, Chinese Academy of Sciences (CASIA)
}
\affil[3]{
	 School of Artificial Intelligence\\
	 University of Chinese Academy of Sciences (UCAS)
}
\affil[4]{
	Center for Excellence in Brain Science and Intelligence Technology (CEBSIT)
}
\affil[5]{
	Chinese Academy of Sciences, Artificial Intelligence Research (CAS-AIR)
}

\begin{document}

\maketitle

\begin{abstract}
	
	Previous methods decompose blind super resolution (SR) problem into two sequential steps: \textit{i}) estimating blur kernel from given low-resolution (LR) image and \textit{ii}) restoring SR image based on estimated kernel. This two-step solution involves two independently trained models, which may not be well compatible with each other. Small estimation error of the first step could cause severe performance drop of the second one. While on the other hand, the first step can only utilize limited information from LR image, which makes it difficult to predict highly accurate blur kernel. Towards these issues, instead of considering these two steps separately, we adopt an alternating optimization algorithm, which can estimate blur kernel and restore SR image in a single model. Specifically, we design two convolutional neural modules, namely \textit{Restorer}  and \textit{Estimator}. \textit{Restorer} restores SR image based on predicted kernel, and \textit{Estimator}  estimates blur kernel with the help of restored SR image. We alternate these two modules repeatedly and unfold this process to form an end-to-end trainable network. In this way, \textit{Estimator} utilizes information from both LR and SR images, which makes the estimation of blur kernel easier. More importantly, \textit{Restorer} is trained with the kernel estimated by \textit{Estimator}, instead of ground-truth kernel, thus \textit{Restorer} could be more tolerant to the estimation error of \textit{Estimator}. Extensive experiments on synthetic datasets and real-world images show that our model can largely outperform state-of-the-art methods and produce more visually favorable results at much higher speed. The source code is available at \url{https://github.com/greatlog/DAN.git}. 
	
\end{abstract}

\section{Introduction}

Single image super resolution (SISR) aims to recover the high-resolution (HR) version of a given degraded low-resolution (LR) image. It has wide applications in video enhancement, medical imaging, as well as security and surveillance imaging. Mathematically, the degradation process can be expressed as
\begin{equation}
\mathbf{y} = (\mathbf{x}\otimes \mathbf{k})\downarrow_{s} + \mathbf{n} \label{downsample}
\end{equation}
where $\mathbf{x}$ is the original HR image, $\mathbf{y}$ is the degraded LR image, $\otimes$ denotes the two-dimensional convolution of $\mathbf{x}$ with blur kernel $\mathbf{k}$, $\mathbf{n}$ denotes Additive White Gaussian Noise (AWGN), and $\downarrow_{s}$ denotes the standard $s$-fold downsampler, which means keeping only the upper-left pixel for each distinct $s\times s$ patch~\cite{usr}. Then SISR refers to the process of recovering $\mathbf{x}$ from $\mathbf{y}$. It is a highly ill-posed problem due to this inverse property, and thus has always been a challenging task.

Recently, deep neural networks (DNNs) have achieved remarkable results on SISR. But most of these methods \cite{rcan,carn,rdn,edsr,san,vdsr} assume that the blur kernel is predefined as the kernel of bicubic interpolation. In this way, large number of training samples can be manually synthesized and further used to train powerful DNNs. However, blur kernels in real applications are much more complicated, and there is a domain gap between bicubically synthesized training samples and real images. This domain gap will lead to severe performance drop when these networks are applied to real applications. Thus, more attention should be paid to SR in the context of unknown blur kernels, \ie blind SR. 

In blind SR, there is one more undetermined variable, \ie blur kernel $\mathbf{k}$, and the optimization also becomes much more difficult. To make this problem easier to be solved, previous methods~\cite{srmd,udvd,dpsr,usr} usually decompose it into two sequential steps: \textit{i}) estimating blur kernel from LR image and \textit{ii}) restoring SR image based on estimated kernel. This two-step solution involves two independently trained models, thus they may be not well compatible to each other. Small estimation error of the first step could cause severe performance drop of the following one~\cite{ikc}. But on the other hand, the first step can only utilize limited information from LR image, which makes it difficult to predict highly accurate blur kernel. As a result, although both models can perform well individually, the final result may be suboptimal when they are combined together. 

Instead of considering these two steps separately, we adopt an alternating optimization algorithm, which can estimate blur kernel $\mathbf{k}$ and restore SR image $\mathbf{x}$ in the same model. Specifically, we design two convolutional neural modules, namely \textit{Restorer} and \textit{Estimator}. \textit{Restorer} restores SR image based on blur kernel predicted by \textit{Estimator}, and the restored SR image is further used to help \textit{Estimator} estimate better blur kernel. Once the blur kernel is manually initialized, the two modules can well corporate with each other to form a closed loop, which can be iterated over and over. The iterating process is then unfolded to an end-to-end trainable network, which is called deep alternating network (DAN). In this way, \textit{Estimator} can utilize information from both LR and SR images, which makes the estimation of blur kernel easier. More importantly, \textit{Restorer} is trained with the kernel estimated by \textit{Estimator}, instead of ground-truth kernel. Thus during testing \textit{Restorer} could be more tolerant to the estimation error of \textit{Estimator}. Besides, the results of both modules could be substantially improved during the iterations, thus it is likely for our alternating optimization algorithm to get better final results than the direct two-step solutions. We summarize our contributions into three points:

\begin{itemize}
	\item [1.] We adopt an alternating optimization algorithm to estimate blur kernel and restore SR image for blind SR in a single network (DAN),  which helps the two modules to be well compatible with each other and likely to get better final results than previous two-step solutions. 

	\item [2.] We design two convolutional neural modules, which can be alternated repeatedly and then unfolded to form an end-to-end trainable network, without any pre/post-processing. It is easier to be trained and has higher speed than previous two-step solutions. To the best of our knowledge, the proposed method is the first end-to-end network for blind SR.
	
	\item [3.] Extensive experiments on synthetic datasets and real-world images show that our model can largely outperform state-of-the art methods and produce more visually favorable results at much higher speed.
\end{itemize}
 
\section{Related Work}
\vspace{-0.1cm}
\subsection{Super Resolution in the Context of Bicubic Interpolation}
\vspace{-0.1cm}
Learning based methods for SISR usually require a large number of paired HR and LR images as training samples. However, these paired samples are hard to get in real world. As a result, researchers manually synthesize LR images from HR images with predefined downsampling settings. The most popular setting is bicubic interpolation, \ie defining $\mathbf{k}$ in Equation~\ref{downsample} as bicubic kernel. From the arising of SRCNN~\cite{srcnn}, various DNNs~\cite{vdsr,rdn,rcan,fsrcnn,dbpn,meta_sr} have been proposed based on this setting. Recently, after the proposal of RCAN~\cite{rcan} and RRDB~\cite{esrgan}, the performance of these non-blind methods even start to saturate on common benchmark datasets. However, the blur kernels for real images are indeed much more complicated. In real applications, kernels are unknown and differ from image to image. As a result, despite that these methods have excellent performance in the context of bicubic downsampling, they still cannot be directly applied to real images due to the domain gap. 

\vspace{-0.1cm}
\subsection{Super Resolution for Multiple Degradations}\label{non-blind}
\vspace{-0.1cm}
Another kind of non-blind SR methods aims to propose a single model for multiple degradations, \ie the second step of the two-step solution for blind SR. These methods take both LR image and its corresponding blur kernel as inputs. In~\cite{zssr_pre,zssr}, the blur kernel is used to downsample images and synthesize training samples, which can be used to train a specific model for given kernel and LR image. In~\cite{srmd}, the blur kernel and LR image are directly concatenated at the first layer of a DNN. Thus, the SR result can be closely correlated to both LR image and blur kernel. In~\cite{dpsr}, Zhang \etal proposed a method based on ADMM algorithm. They interpret this problem as MAP optimization and solve the data term and prior term alternately. In~\cite{ikc}, a spatial feature transform (SFT) layer is proposed to better preserve the details in LR image while blur kernel is an additional input. However, as pointed out in~\cite{ikc}, the SR results of these methods are usually sensitive to the provided blur kernels. Small deviation of provided kernel from the ground truth will cause severe performance drop of these non-blind SR methods.
\vspace{-0.1cm}
\subsection{Blind Super Resolution}
\vspace{-0.1cm}
Previous methods for blind SR are usually the sequential combinations of a kernel-estimation method and a non-blind SR method. Thus kernel-estimation methods are also an important part of blind SR. In~\cite{nonpara}, Michaeli \etal estimate the blur kernel by utilizing the internal patch recurrence. In~\cite{kernel_gan} and~\cite{gan_first}, LR image is firstly downsampled by a generative network, and then a discriminator is used to verify whether the downsampled image has the same distribution with original LR image. In this way, the blur kernel can be learned by the generative network. In~\cite{ikc}, Gu \etal not only train a network for kernel estimation, but also propose a correction network to iteratively correct the kernel. Although the accuracy of estimated kernel is largely improved, it requires training of two or even three networks, which is rather complicated. Instead, DAN is an end-to-end trainable network that is much easier to be trained and has much higher speed.

\begin{figure}
	\centering
	\includegraphics[width=\linewidth]{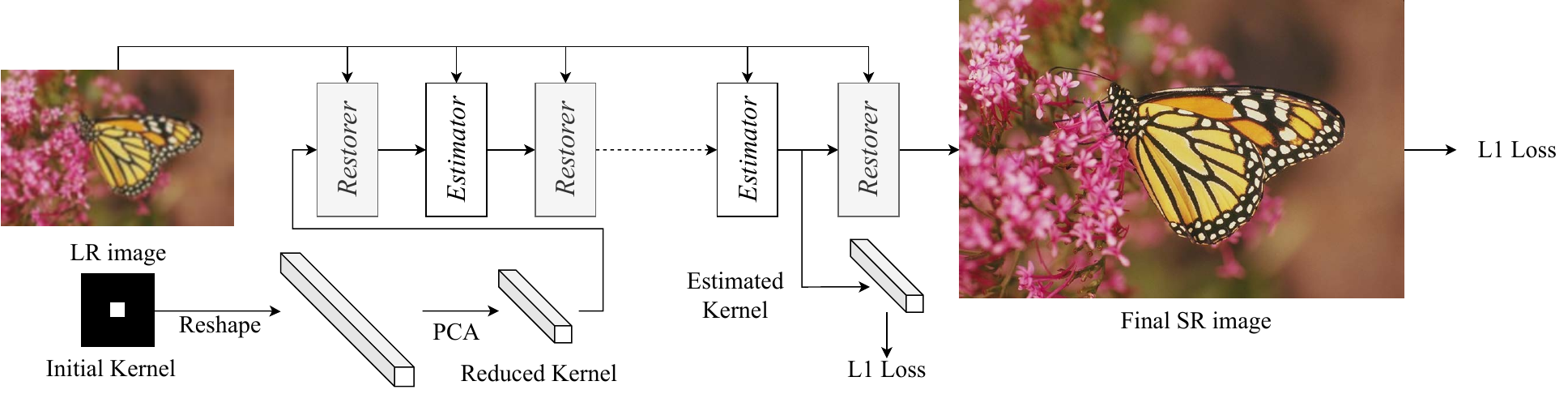}
	\caption{The overview structure of the deep alternating network (DAN).} \label{overview}
	\vspace{-0.2cm}
\end{figure}

\section{End-to-End Blind Super Resolution}
\subsection{Problem Formulation}\label{formulation}
As shown in Equation~\ref{downsample}, there are three variables, \ie $\mathbf{x}$, $\mathbf{k}$ and $\mathbf{n}$, to be determined in blind SR problem. In literature, we can apply a denoise algorithm~\cite{ircan,bm3d,wnnm} in the first place. Then blind SR algorithm only needs to focus on solving $\mathbf{k}$ and $\mathbf{x}$. It can be mathematically expressed an optimization problem: 
\begin{equation}
\mathop{\arg\min}_{\mathbf{k},\mathbf{x}}  \|\mathbf{y} - (\mathbf{x}\otimes \mathbf{k})\downarrow_{s} \|_1 + \phi(\mathbf{x})
\end{equation}
where the former part is the reconstruction term, and $\phi(\mathbf{x})$ is prior term for HR image. The prior term is usually unknown and has no analytic expression. Thus it is extremely difficult to solve this problem directly. Previous methods decompose this problem into two sequential steps:
\begin{equation}
\left \{
\begin{aligned}
\mathbf{k} &= M(\mathbf{y})\\
\mathbf{x} &= \mathop{\arg\min}_{\mathbf{x}}  \|\mathbf{y} - (\mathbf{x}\otimes \mathbf{k})\downarrow_{s} \|_1 + \phi(\mathbf{x})
\end{aligned}
\right.
\end{equation}
where $M(\cdot)$ denotes the function that estimates $\mathbf{k}$ from $\mathbf{y}$, and the second step is usually solved by a non-blind SR method described in Sec~\ref{non-blind}. This two-step solution has its drawbacks in threefold. Firstly, this algorithm usually requires training of two or even more models, which is rather complicated. Secondly, $M(\cdot)$ can only utilize information from $\mathbf{y}$, which treats $\mathbf{k}$ as a kind of prior of $\mathbf{y}$. But in fact, $\mathbf{k}$ could not be properly solved without information from $\mathbf{x}$. At last, the non-blind SR model for the second step is trained with ground-truth kernels. While during testing, it can only have access to kernels estimated in the first step. The difference between ground-truth and estimated kernels will usually cause serve performance drop of the non-blind SR model~\cite{ikc}. 

Towards these drawbacks, we propose an end-to-end network that can largely release these issues. We still split it into two subproblems, but instead of solving them sequentially, we adopt an alternating optimization algorithm, which restores SR image and estimates corresponding blur kernel alternately. The mathematical expression is
\begin{equation}
\left \{
\begin{aligned}
\mathbf{k}_{i+1} &=  \mathop{\arg\min}_{\mathbf{k}}  \|\mathbf{y} - (\mathbf{x}_{i}\otimes \mathbf{k})\downarrow_{s} \|_1\\
\mathbf{x}_{i+1} &= \mathop{\arg\min}_{\mathbf{x}}  \|\mathbf{y} - (\mathbf{x}\otimes \mathbf{k}_{i})\downarrow_{s} \|_1 + \phi(\mathbf{x})
\end{aligned}
\right.
\end{equation}
We alternately solve these two subproblems both via convolutional neural modules, namely \textit{Estimator} and \textit{Restorer} respectively. Actually, there even has an analytic solution for \textit{Estimator}. But we experimentally find that analytic solution is more time-consuming and not robust enough (when noise is not fully removed). We fix the number of iterations as $T$ and unfold the iterating process to form an end-to-end trainable network, which is called deep alternating network (DAN). 

As shown in Figure~\ref{overview}, we initialize the kernel by Dirac function, \ie the center of the kernel is one and zeros otherwise. Following~\cite{ikc,srmd}, the kernel is also reshaped and then reduced by principal component analysis (PCA). We set $T=4$ in practice and both modules are supervised only at the last iteration by L1 loss. The whole network could be well trained without any restrictions on intermediate results, because the parameters of both modules are shared between different iterations. 

In DAN, \textit{Estimator} takes both LR and SR images as inputs, which makes the estimation of blur kernel $\mathbf{k}$ much easier. More importantly, \textit{Restorer} is trained with the kernel estimated by \textit{Estimator}, instead of ground-truth kernel as previous methods do. Thus, \textit{Restorer} could be more tolerant to the estimation error of \textit{Estimator} during testing. Besides, compared with previous two-step solutions, the results of both modules in DAN could be substantially improved, and it is likely for DAN to get better final results. Specially, in the case where the scale factor $s=1$, DAN becomes an deblurring network. Due to limited pages, we only discuss SR cases in this paper.

\subsection{Instantiate the Convolutional Neural Modules }

Both modules in our network have two inputs. \textit{Estimator} takes LR and SR image, and \textit{Restorer} takes LR image and blur kernel as inputs. We define LR image as basic input, and the other one is conditional input. For example, blur kernel is the conditional input of \textit{Restorer}. During iterations, the basic inputs of both modules keep the same, but their conditional inputs are repeatedly updated. We claim that it is significantly important to keep the output of each module closely related to its conditional input. Otherwise, the iterating results will collapse to a fixed point at the first iteration. Specifically, if \textit{Estimator} outputs the same kernel regardless the value of SR image, or \textit{Restorer} outputs the same SR image regardless of the value of blur kernel, their outputs will only depend on the basic input, and the results will keep the same during the iterations.

\begin{figure}[t]
	\centering
	\includegraphics[width=\linewidth]{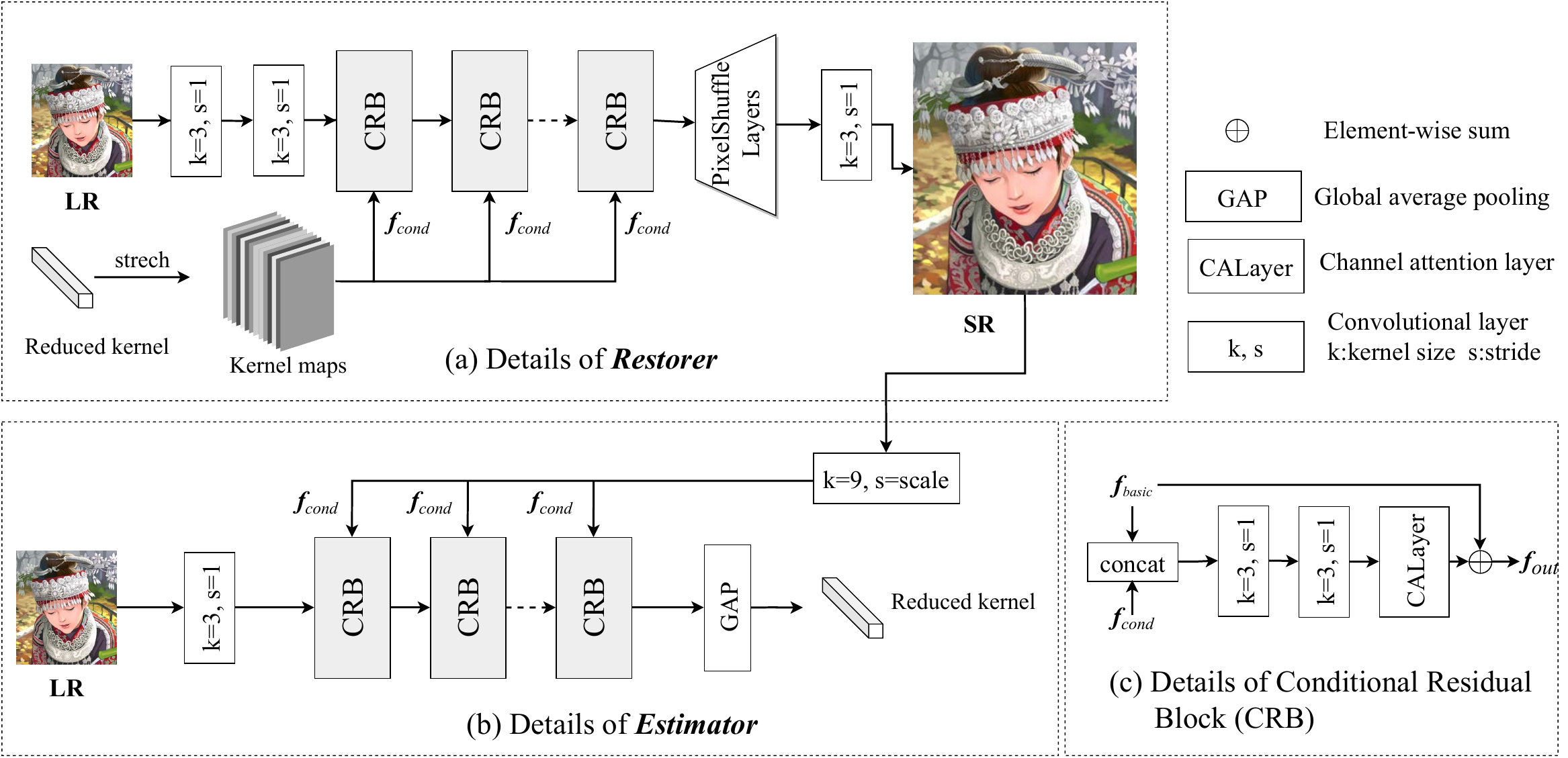}
	\caption{The details of \textit{Estimator}  and \textit{Restorer}. (a) (top) The details of \textit{Restorer}. (c) (bottom-left) The details of \textit{Estimator}. (c) (bottom-right) The details of conditional residual block (CRB).} \label{modules}
	\vspace{-0.5cm}
\end{figure}

To ensure that the outputs of \textit{Estimator} and \textit{Restorer} are closely related to their conditional inputs, we propose a conditional residual block (CRB). On the basis of the residual block in~\cite{rcan}, we concatenate the conditional and basic inputs at the beginning:
\begin{equation}
	f_{out} = R(Concat([f_{basic}, f_{cond}])) + f_{basic}
\end{equation}
where $R(\cdot)$ denotes the residual mapping function of CRB and $Concat([\cdot, \cdot])$ denotes concatenation. $f_{basic}$ and $f_{cond}$ are the basic input and conditional input respectively. As shown in Figure~\ref{modules} (c), the residual mapping function consists of two $3\times3$ convolutional layers and one channel attention layer~\cite{senet}. Both \textit{Estimator} and \textit{Restorer} are build by CRBs.

\textbf{\textit{Estimator}}. The whole structure of \textit{Estimator} is shown in Figure~\ref{modules} (b). We firstly downsample SR image by a  convolutional layer with stride $s$. Then the feature maps are sent to all CRBs as conditional inputs. At the end of the network, we squeeze the features by global average pooling to form the elements of predicted kernel. Since the kernel is reduced by PCA, \textit{Estimator} only needs to estimate the PCA result of blur kernel. In practice, \textit{Estimator} has $5$ CRBs, and both basic input and conditional input of  each CRB have $32$ channels. 

\textbf{\textit{Restorer}}. The whole structure of \textit{Restorer} is shown in Figure~\ref{modules} (a). In \textit{Restorer}, we stretch the kernel in spatial dimension to the same spatial size as LR image. Then the stretched kernel is sent to all CRBs of \textit{Restorer} as conditional inputs. We use PixelShuffle~\cite{pixel_shuffle} layers to upscale the features to desired size. In practice, \textit{Restorer} has $40$ CRBs, and the basic input and conditional input of each CRB has $64$ and $10$ channels respectively.

\section {Experiments} 

\subsection{Experiments on Synthetic Test Images}

\subsubsection{Data, Training and Testing}

We collect $3450$ HR images from DIV2K~\cite{div2k} and Flickr2K~\cite{flickr2k} as training set. To make reasonable comparison with other methods, we train models with two different degradation settings. One is the setting in~\cite{ikc}, which only focuses on cases with isotropic Gaussian blur kernels. The other is the setting in~\cite{kernel_gan}, which focuses on cases with more general and irregular blur kernels.  

\textbf{Setting 1.} Following the setting in~\cite{ikc}, the kernel size is set as $21$. During training,  the kernel width is uniformly sampled in [0.2, 4.0], [0.2, 3.0] and [0.2, 2.0] for scale factors $4$, $3$ and $2$ respectively. For quantitative evaluation, we collect HR images from the commonly used benchmark datasets, \ie Set5~\cite{set5}, Set14~\cite{set14}, Urban100~\cite{urban100}, BSD100~\cite{bsd100} and Manga109~\cite{manga109}. Since determined kernels are needed for reasonable comparison, we uniformly choose 8 kernels, denoted as \textit{Gaussian8}, from range [1.8, 3.2], [1.35, 2.40] and [0.80, 1.60] for scale factors $4$, $3$ and $2$ respectively. The HR images are first blurred by the selected blur kernels and then downsampled to form synthetic test images. 

\textbf{Setting 2.} Following the setting in~\cite{kernel_gan},  we set the kernel size as $11$. We firstly generate anisotropic Gaussian kernels. The lengths of both axises are uniformly distributed in $(0.6, 5)$, rotated by a random angle uniformly distributed in [$-\pi$, $\pi$]. To deviate from a regular Gaussian, we further apply uniform multiplicative noise (up to 25\% of each pixel value of the kernel) and normalize it to sum to one. For testing, we use the benchmark dataset DIV2KRK that is used in~\cite{kernel_gan}.

The input size during training is $64\times64$ for all scale factors. The batch size is $64$. Each model is trained for $4\times10^{5}$ iterations. We use Adam~\cite{adam} as our optimizer, with $\beta_1=0.9$, $\beta_2=0.99$. The initial learning rate is $2\times10^{-4}$, and will decay by half at every $1\times10^5$ iterations. All models are trained on RTX2080Ti GPUs.

\subsubsection{Quantitative Results}

\textbf{Setting 1.} For the first setting, we evaluate our method on test images synthesized by \textit{Gaussian8} kernels. We mainly compare our results with ZSSR~\cite{zssr} and IKC~\cite{ikc}, which are methods designed for blind SR. We also include a comparison with CARN~\cite{carn}. Since it is not designed for blind SR, we perform deblurring method~\cite{pan} before or after CARN. The PSNR and SSIM results on Y channel of transformed YCbCr space are shown in Table~\ref{setting1}. 

Despite that CARN achieves remarkable results in the context of bicubic downsampling, it suffers severe performance drop when applied to images with unknown blur kernels. Its performance is largely improved when it is followed by a deblurring method,  but still inferior to that of blind-SR methods. ZSSR trains specific network for each single tested image by utilizing the internal patch recurrence. However, ZSSR has an in-born drawback: the training samples for each image are limited, and thus it cannot learn a good prior for HR images. IKC is also a two-step solution for blind SR. Although the accuracy of estimated kernel is largely improved in IKC, the final result is still suboptimal. DAN is trained in an end-to-end scheme, which is not only much easier to be trained than two-step solutions, but also likely to a reach a better optimum point. As shown in Table~\ref{setting1}, the PSNR result of DAN on Manga109 for scale $\times3$ is even $4.95dB$ higher than that of IKC. For other scales and datasets, DAN also largely outperforms IKC. 

\begin{figure}[t]
	\vspace{-0.3cm}
	\centering
	\includegraphics[width=\linewidth]{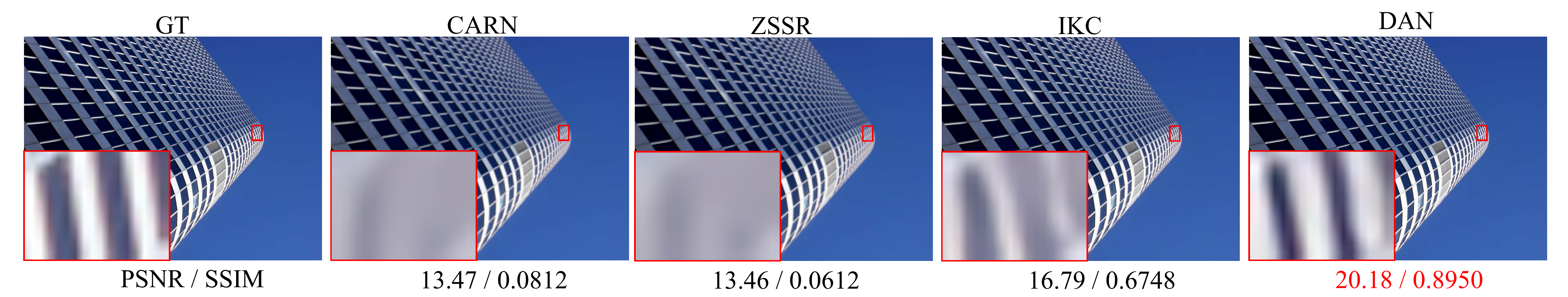}
	\vspace{-0.6cm}
	\caption{Visual results of \textit{img 005} in Urban100. The width of blur kernel is $1.8$.}\label{visual_setting1}
	\vspace{-0.5cm}
\end{figure}

The visual results of \textit{img 005} in Urban100 are shown in Figure~\ref{visual_setting1} for comparison. As one can see, CARN and ZSSR even cannot restore the edges for the window. IKC performs better, but the edges are severely blurred. While DAN can restore sharp edges and produce more visually pleasant result.

\begin{table}[t]
	\centering
	\caption{Quantitative comparison with SOTA SR methods with Setting 1. The best two results are highlighted in red and blue colors respectively.} \label{setting1}
	\resizebox{\linewidth}{!}{
		\begin{tabular}{cccccccccccc} 
			\hline
			\multirow{2}{*}{Method} & \multirow{2}{*}{Scale} & \multicolumn{2}{c}{Set5} & \multicolumn{2}{c}{Set14} & \multicolumn{2}{c}{BSD100} & \multicolumn{2}{c}{Urban100} & \multicolumn{2}{c}{Manga109} \\
			&                        & ~~PSNR~~       & ~~SSIM~~        & ~~PSNR~~        & ~~SSIM~~        & ~~PSNR~~        & ~~SSIM~~         & ~~PSNR~~         & ~~SSIM~~          & ~~PSNR~~         & ~~SSIM~~          \\ 
			\hline
			\hline
			Bicubic                 & \multirow{7}{*}{2}     & 28.82      & 0.8577      & 26.02       & 0.7634      & 25.92       & 0.7310       & 23.14        & 0.7258        & 25.60        & 0.8498        \\
			CARN~\cite{carn}                    &                        & 30.99      & 0.8779      & 28.10       & 0.7879      & 26.78       & 0.7286       & 25.27        & 0.7630        & 26.86        & 0.8606        \\
			ZSSR~\cite{zssr}                    &                        & 31.08      & 0.8786      & 28.35       & 0.7933      & 27.92       & 0.7632       & 25.25        & 0.7618        & 28.05        & 0.8769        \\
			\cite{pan}+CARN~\cite{carn}                 &                        & 24.20      & 0.7496      & 21.12       & 0.6170      & 22.69       & 0.6471       & 18.89        & 0.5895        & 21.54        & 0.7946        \\
			CARN~\cite{carn}+\cite{pan}                  &                        & 31.27      & 0.8974      & 29.03       & 0.8267      & 28.72       & 0.8033       & 25.62        & 0.7981        & 29.58        & 0.9134        \\
			IKC~\cite{ikc}                     &                        &\color{blue}36.62      &\color{blue}0.9658      &\color{blue}32.82       &\color{blue}0.8999      &\color{blue}31.36       &\color{blue}0.9097       &\color{blue}30.36        &\color{blue}0.8949        &\color{blue}36.06        &\color{blue}0.9474        \\
			DAN                             &                    &\color{red}37.33         &\color{red}0.9754            &\color{red}33.07         &\color{red}0.9068            &\color{red}31.76           &\color{red}0.9213            &\color{red}30.60              &\color{red}0.9020               &\color{red}37.23             &\color{red}0.9710             \\ 
			\hline
			Bicubic                 & \multirow{7}{*}{3}     & 26.21      & 0.7766      & 24.01       & 0.6662      & 24.25       & 0.6356       & 21.39        & 0.6203        & 22.98        & 0.7576        \\
			CARN~\cite{carn}                    &                        & 27.26      & 0.7855      & 25.06       & 0.6676      & 25.85       & 0.6566       & 22.67        & 0.6323        & 23.84        & 0.7620        \\
			ZSSR~\cite{zssr}                    &                        & 28.25      & 0.7989      & 26.11       & 0.6942      & 26.06       & 0.6633       & 23.26        & 0.6534        & 25.19        & 0.7914        \\
			\cite{pan}+CARN~\cite{carn}                  &                        & 19.05      & 0.5226      & 17.61       & 0.4558      & 20.51       & 0.5331       & 16.72        & 0.4578        & 18.38        & 0.6118        \\
			CARN~\cite{carn}+\cite{pan}                &                        & 30.31      & 0.8562      & 2757        & 0.7531      & 27.14       & 0.7152       & 24.45        & 0.7241        & 27.67        & 0.8592        \\
			IKC~\cite{ikc}                     &                        &\color{blue}32.16      &\color{blue}0.9420      &\color{blue}29.46       &\color{blue}0.8229      &\color{blue}28.56       &\color{blue}0.8493       &\color{blue}25.94        &\color{blue}0.8165        &\color{blue}28.21        &\color{blue}0.8739        \\
			DAN                            &                        &\color{red}34.04            &\color{red}0.9620           &\color{red}30.09             &\color{red}0.8287           &\color{red}28.94            &\color{red}0.8606              &\color{red}27.65              &\color{red}0.8352              &\color{red}33.16              &\color{red}0.9382               \\ 
			\hline
			Bicubic                 & \multirow{7}{*}{4}     & 24.57      & 0.7108      & 22.79       & 0.6032      & 23.29       & 0.5786       & 20.35        & 0.5532        & 21.50        & 0.6933        \\
			CARN~\cite{carn}                    &                        & 26.57      & 0.7420      & 24.62       & 0.6226      & 24.79       & 0.5963       & 22.17        & 0.5865        & 21.85        & 0.6834        \\
			ZSSR~\cite{zssr}                    &                        & 26.45      & 0.7279      & 24.78       & 0.6268      & 24.97       & 0.5989       & 22.11        & 0.5805        & 23.53        & 0.7240        \\
			\cite{pan}+CARN~\cite{carn}                  &                        & 18.10      & 0.4843      & 16.59       & 0.3994      & 18.46       & 0.4481       & 15.47        & 0.3872        & 16.78        & 0.5371        \\
			CARN~\cite{carn}+\cite{pan}                 &                        & 28.69      & 0.8092      & 26.40       & 0.6926      & 26.10       & 0.6528       & 23.46        & 0.6597        & 25.84        & 0.8035        \\
			IKC~\cite{ikc}                    &                        &\color{blue}31.52      &\color{blue}0.9278      &\color{blue}28.26       &\color{blue}0.7688      & \color{blue}27.29       &\color{blue}0.8041       &\color{blue}25.33        &\color{blue}0.7760        &\color{blue}29.90        &\color{blue}0.8793        \\
			DAN              &                        &\color{red}31.89        &\color{red}0.9302      &\color{red}28.43       &\color{red}0.7693      & \color{red}27.51        &\color{red}0.8078      &\color{red}25.86     &\color{red}0.7822 &\color{red}30.50& \color{red}0.9037             \\ 
			\hline
	\end{tabular}}
\end{table}

\textbf{Setting 2.} The second setting involves irregular blur kernels, which is more general, but also more difficult to solve. For Setting 2, we mainly compare methods of three different classes: \textit{i}) SOTA SR algorithms trained on bicubically downsampled images such as EDSR~\cite{edsr} and RCAN~\cite{rcan} , \textit{ii}) blind SR methods designed for NTIRE competition such as PDN~\cite{pdn} and WDSR~\cite{wdsr}, \textit{iii}) the two-step solutions, \ie the combination of a kernel estimation method and a non-blind SR method, such as Kernel-GAN~\cite{kernel_gan} and ZSSR~\cite{zssr}. The PSNR and SSIM results on Y channl are shown in Table~\ref{setting2}. 

Similarly, the performance of methods trained on bicubically downsampled images is limited by the domain gap. Thus, their results are only slightly better than that of interpolation. The methods in Class 2 are trained on synthesized images provided in NTIRE competition. Although these methods achieve remarkable results in the competition, they still cannot generalize well to irregular blur kernels. 

The comparison between methods of Class 3 can enlighten us a lot. Specifically, USRNet~\cite{usr} achieves remarkable results when GT kernels are provided, and KernelGAN also performs well on kernel estimation. However, when they are combined together, as shown in Table~\ref{setting2}, the final SR results are worse than all other methods. This indicates that it is important for the \textit{Estimator} and \textit{Restorer} to be compatible with each other. Additionally, although better kernel-estimation method can benefit the SR results, the overall performance is still largely inferior to that of DAN. DAN outperforms the combination of KernelGAN and ZSSR by $2.20dB$ and $0.74dB$ for scales $\times2$ and $\times4$ respectively. 

\begin{figure}
	\centering
	\includegraphics[width=\linewidth]{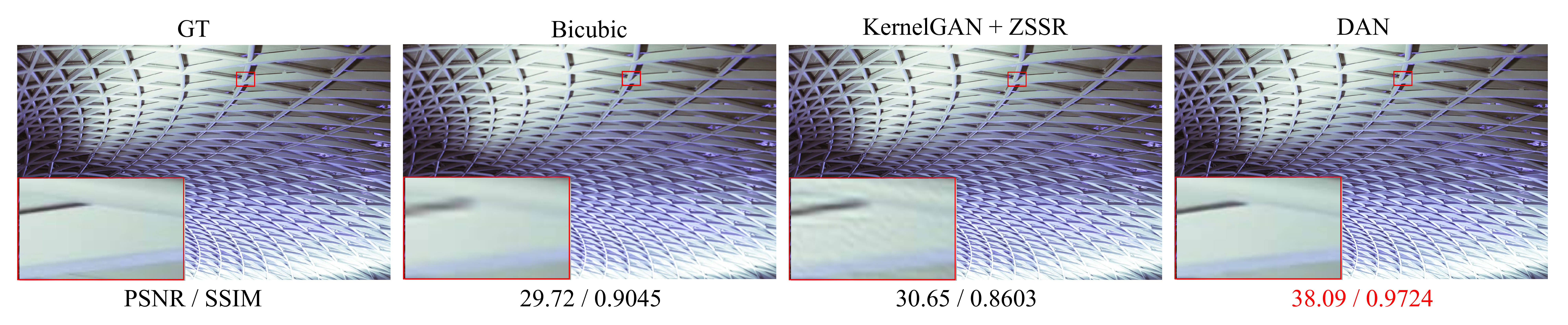}
	\vspace{-0.5cm}
	\caption{Visual results on \textit{img 892} in DIV2KRK.} \label{visual_setting2}
	\vspace{-0.5cm}
\end{figure}

The visual results of \textit{img 892} in DIVKRK are shown in Figure~\ref{visual_setting2}. Although the combination of KernelGAN and ZSSR can produce slightly shaper edges than interpolation, it suffers from severe artifacts. The SR image of DAN is obviously much cleaner and has more reliable details.

\begin{table}[t]
	\centering
	\caption{Quantitative comparison with SOTA SR methods with Setting 2. The best two results are highlighted in red and blue colors respectively.} \label{setting2}
	\footnotesize 
	\begin{tabular}{cccccc}
		\hline
		\multirow{3}{*}{Types}   
		& \multirow{3}{*}{Method}     & \multicolumn{4}{c}{Scale}                     \\
		&                             & \multicolumn{2}{c}{2} & \multicolumn{2}{c}{4} \\
		&                             & ~~PSNR~~      & ~~SSIM~~      & ~~PSNR~~      & ~~SSIM~~      \\ 
		\hline
		\hline
		\multirow{4}{*}{Class 1}
		& Bicubic                     & 28.73     & 0.8040    & 25.33     & 0.6795    \\
		& Bicubic kernel + ZSSR~\cite{zssr}       & 29.10     & 0.8215    & 25.61     & 0.6911    \\
		& EDSR~\cite{edsr}                    & 29.17     & 0.8216    & 25.64     & 0.6928    \\
		& RCAN~\cite{rcan}                  & 29.20     & 0.8223    & 25.66     & 0.6936    \\ \hline
		\multirow{5}{*}{Class 2} 
		&~~PDN~\cite{pdn} - 1st in NTIRE'19 track4             & /   & /     & 26.34     & 0.7190    \\
		& ~~WDSR~\cite{wdsr} - 1st in NTIIRE'19 track2~~   & /   & /     & 21.55     & 0.6841    \\
		& WDSR~\cite{wdsr} - 1st in NTIRE'19 track3            & /   & /      & 21.54     & 0.7016    \\
		&~~WDSR~\cite{wdsr} - 2nd in NTIRE'19 track4~~    &  /    & /       & 25.64     & 0.7144    \\ 
		&Ji \etal ~\cite{ji2020real} - 1st in NITRE'20 track 1 &  /  & /     & 25.43     & 0.6907 \\ \hline
		\multirow{6}{*}{Class 3}
		& Cornillere \etal~\cite{siga}& 29.46     & 0.8474    & /     & /   \\
		& Michaeli \etal~\cite{nonpara} + SRMD ~\cite{srmd}              & 25.51     & 0.8083    & 23.34     & 0.6530    \\
		& Michaeli \etal~\cite{nonpara} + ZSSR~\cite{zssr}               & 29.37     & 0.8370    & 26.09     & 0.7138    \\
		& KernelGAN~\cite{kernel_gan} + SRMD~\cite{srmd}            & 29.57     & 0.8564    & 25.71     & 0.7265    \\
		& KernelGAN~\cite{kernel_gan} + USRNet~\cite{usr}             & /          &   /             & 20.06     & 0.5359 \\
		& KernelGAN ~\cite{kernel_gan}+ ZSSR~\cite{zssr}           &\color{blue}30.36     &\color{blue}0.8669    &\color{blue}26.81     &\color{blue}0.7316    \\ \hline
		Ours                     & DAN                      &\color{red}32.56&\color{red}0.8997          &\color{red} 27.55 &\color {red}0.7582        \\ 
		\hline
	\end{tabular}
\vspace{-0.2cm}
\end{table}

\begin{table}[h]
	\centering
	\caption{PSNR results when GT kernels are provided.}\label{gt_kernel}
	\resizebox{0.8\linewidth}{!}{
		\begin{tabular}{c|cccccc}
			\hline
			Methods & Set5  & Set14 & B100  & Urban100 & Manga109 \\ \hline
			DAN & 31.89 & 28.43 & 27.51 & 25.86    & 30.50    \\
			DAN(GT kernel) & 31.85 & 28.42 & 27.51 & 25.87    & 30.51    \\ \hline
		\end{tabular}
	}
\end{table}

\subsubsection{Study of Estimated Kernels}

\begin{figure}
	\centering
	\subfigure[Kernel loss on different sigma.]{
		\begin{minipage}{0.42\linewidth}
			\includegraphics[width=\linewidth]{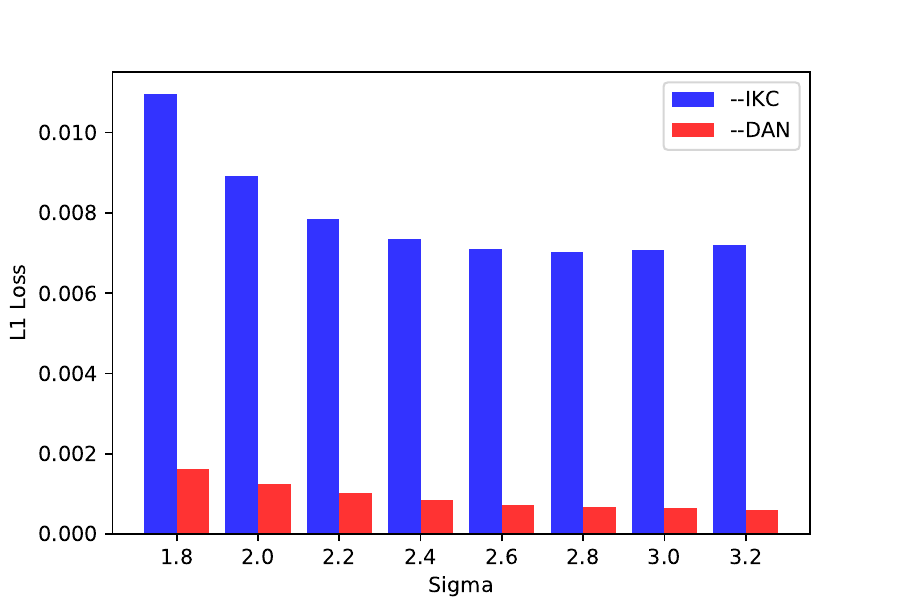}
		\end{minipage}
	}
	\subfigure[PSNR over kernels with  different sigma.]{
		\begin{minipage}{0.42\linewidth}
			\includegraphics[width=\linewidth]{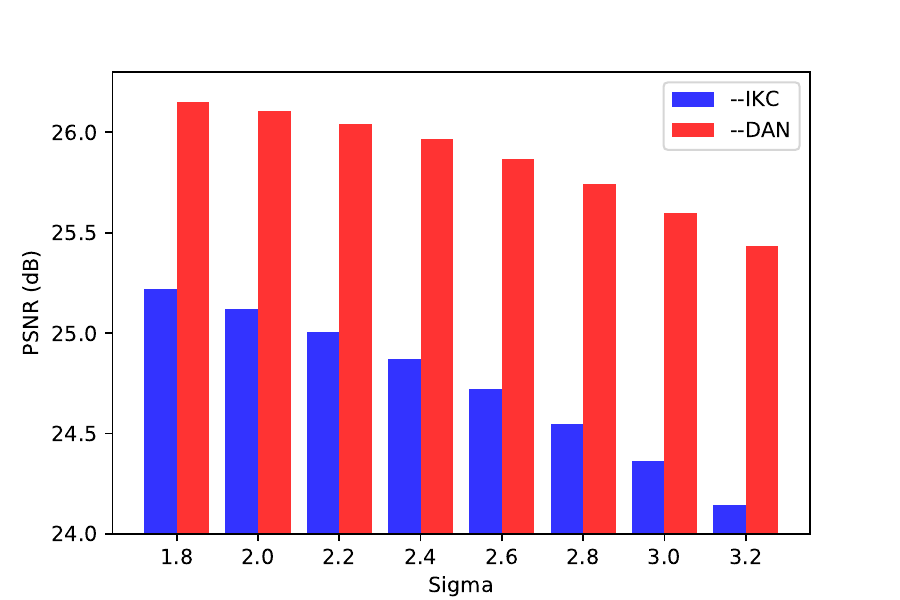}
		\end{minipage}
	}
\caption{The L1 loss of predicted kernels with different sigma (left) and PSNR results with respect to different kernels (right).}\label{kernel_results}
\end{figure}

To evaluated the accuracy of predicted kernels, we calculate their L1 errors in the reduced space, and the results on Urban100 are shown in Figure~\ref{kernel_results} (a). As one can see that the L1 error of reduced kernels predicted by DAN are much lower than that of IKC. It suggests that the overall improvements of DAN may partially come from more accurate retrieved kernels.  We also plot the PSNR results with respect to kernels with different sigma in Figure~\ref{kernel_results} (b). As sigma increases, the performance gap between IKC and DAN also becomes larger. It indicates that DAN may have better generalization ability.

We also replace the estimated kernel by ground truth (GT) to further investigate the influence of \textit{Estimator}. If GT kernels are provided, the iterating processing becomes meaningless. Thus we test the \textit{Restorer} with just once forward propagation. The tested results for Setting 1 is shown in Table~\ref{gt_kernel}. The result almost keeps unchanged and sometimes even gets worser when GT kernels are provided. It indicates that \textit{Predictor} may have already satisfied the requirements of \textit{Restorer}, and the superiority of DAN also partially comes from this good cooperation between its \textit{Predictor} and \textit{Restorer}.

\subsubsection{Study of Iterations}

\begin{figure}[h]
	\centering
	\includegraphics[width=\linewidth]{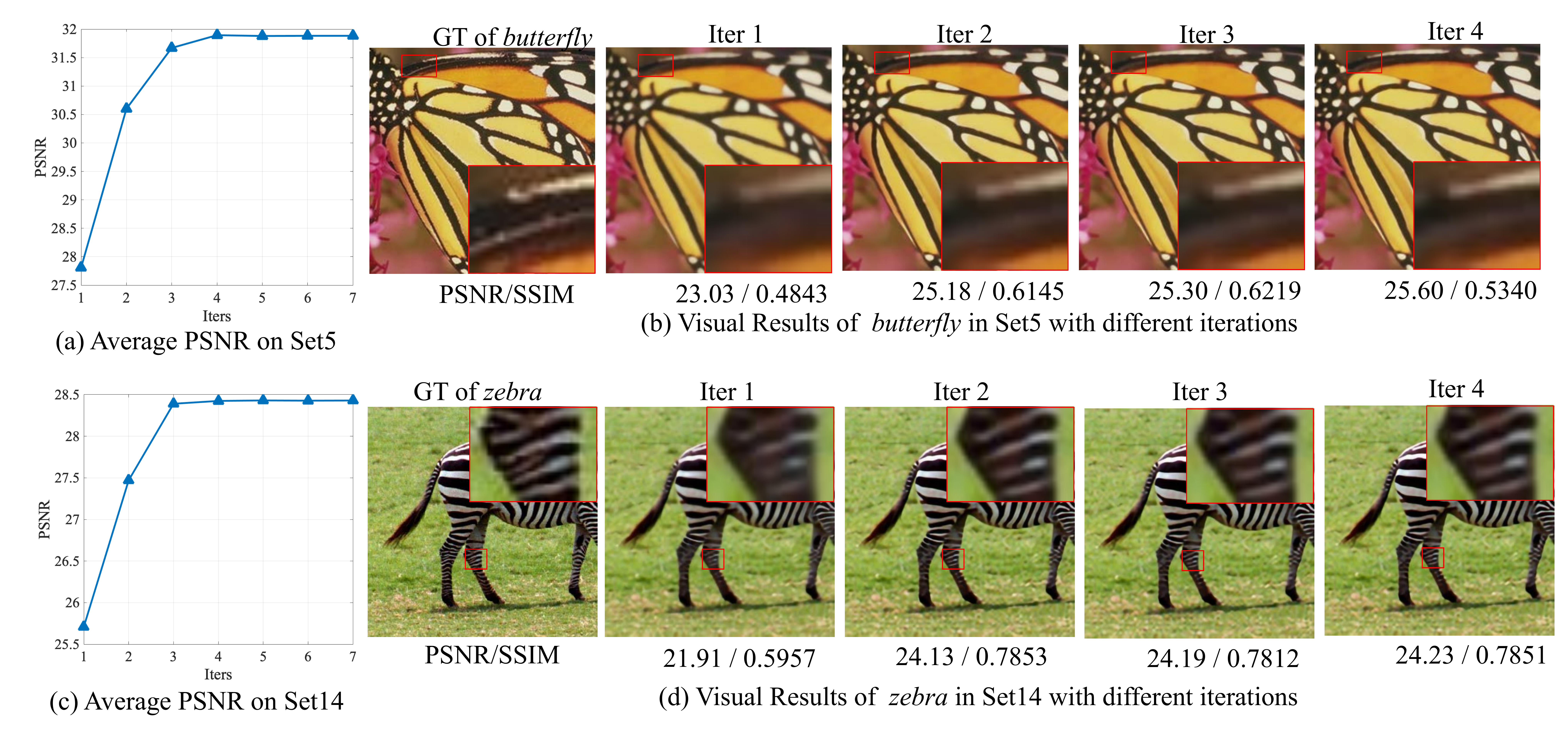}
	\caption{PSNR and visual results with different iterations during testing on Set5 and Set14.}\label{iter}
\end{figure}

After the model is trained, we also change the number of iterations to see whether the two modules have learned the property of convergence or just have `remembered' the iteration number. The model is trained with $4$ iterations, but during testing we increase the iteration number from $1$ to $7$. As shown in Figure~\ref{iter} (a) and (c), the average PSNR results on Set5 and Set14 firstly increase rapidly and then gradually converge. It should be noted that when we iterate more times than training, the performance dose not becomes worse, and sometimes even becomes better. For example, the average PSNR on Set14 is $20.43dB$ when the iteration number is $5$, higher than $20.42dB$ when we iterate $4$ times. Although the incremental is relatively small, it suggests that the two modules may have learned to cooperate with each other, instead of solving this problem like ordinary end-to-end networks, in which cases, the performance will drop significantly when the setting of testing is different from that of training. It also suggests that the estimation error of intermediate results does not destroy the convergence of DAN. In other words, DAN is robust to various estimation error.

\subsection{Inference Speed}

One more superiority of our end-to-end model is that it has higher inference speed. To make a quantitative comparison, we evaluate the average speed of different methods on the same platform. We choose the 40 images synthesized by \textit{Gaussian8}  kernels from Set5 as testing images, and all methods are evaluated on the same platform with a RTX2080Ti GPU. We choose KernelGAN~\cite{kernel_gan} + ZSSR~\cite{zssr} as the one of the representative methods. Its speed is \textbf{415.7} seconds per image. IKC~\cite{ikc} has much faster inference speed, which is only \textbf{3.93} seconds per image. As a comparison, the average speed of DAN is \textbf{0.75} seconds per image,  nearly 554 times faster than KernelGAN + ZSSR, and 5 times faster than IKC. In other words, DAN not only can largely outperform SOTA blind SR methods on PSNR results, but also has much higher speed.

\subsection{Experiments on Real World Images}

We also conduct experiments to prove that DAN can generalize well to real wold images. In this case, we need to consider the influence of additive noise. As we mentioned in Sec~\ref{formulation}, we can perform an denoise algorithm in the first place. But for simplicity, we retrain a different model by adding AWGN to LR image during training. In this way, DAN would be forced to generalize to noisy images. The covariance of noise is set as $15$. We use KernelGAN~\cite{kernel_gan} + ZSSR~\cite{zssr} and IKC~\cite{ikc} as the representative methods for blind SR, and CARN~\cite{carn} as the representative method for non-blind SR method. The commonly used image \textit{chip}~\cite{chip} is chosen as test image. It should be noted that it is a real image and we do not have the ground truth. Thus we can only provide a visual comparison in Figure~\cite{chip}. As one can see, the result of KernelGAN + ZSSR is slightly better than bicubic interpolation, but is still heavily blurred. The result of CARN is over smoothed and the edge is not sharp enough. IKC produces cleaner result, but there are still some artifacts. The letter `X' restored by IKC has an obvious dark line at the top right part. But this dark line is much lighter in the image restored by DAN. It suggests that if trained with noisy images, DAN can also learn to denoise, and produce more visually pleasant results with more reliable details. This is because that both modules are implemented via convolutional layers, which are flexible enough to be adapted to different tasks.  

\begin{figure}[h]
	\centering
	\includegraphics[width=\linewidth]{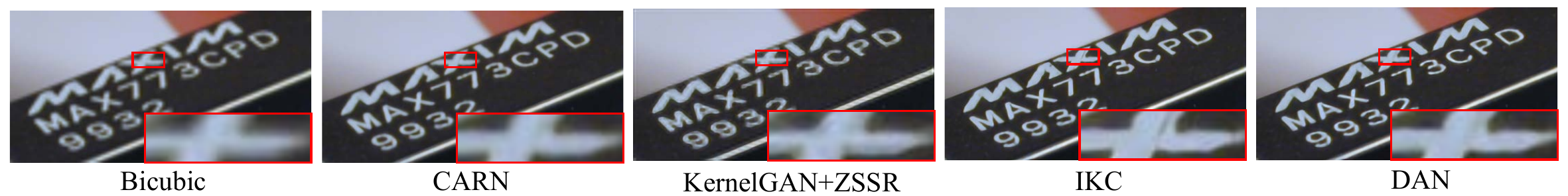}
	\vspace{-0.5cm}
	\caption{Visual results on real image \textit{chip}.}\label{chip}
	\vspace{-0.5cm}
\end{figure}

\section{Conclusion}

In this paper, we have proposed an end-to-end algorithm for blind SR. This algorithm is based on alternating optimization, the two parts of which are both implemented by convolutional modules, namely \textit{Restorer} and \textit{Estimator}. We unfold the alternating process to form an end-to-end trainable network. In this way, \textit{Estimator} can utilize information from both LR and SR images, which makes it easier to estimate blur kernel. More importantly, \textit{Restorer} is trained with the kernel estimated by \textit{Estimator}, instead of ground-truth kernel, thus \textit{Restorer} could be more tolerant to with the estimation error of \textit{Estimator}. Besides, the results of both modules could be substantially improved during the iterations, thus it is likely for DAN to get better final results than previous two-step solutions. Experiments also prove that DAN outperforms SOTA blind SR methods by a large margin. In the future, if the two parts of DAN can be implemented by more powerful modules, we believe that its performance could be further improved.

\section*{ Broader Impact}

Super Resolution is a traditional task in computer vision. It has been studied for several decades and has wide applications in video enhancement, medical imaging, as well as security and surveillance imaging. These techniques have largely benefited the society in various areas for years and have no negative impact yet. The proposed method (DAN) could further improve the merits of these applications especially in cases where the degradations are unknown. DAN has relatively better performance and much higher speed, and it is possible for DAN to be used in real-time video enhancement or surveillance imaging. This work does not present any negative foreseeable societal consequence.

\section*{Acknowledgements}
This work is jointly supported by National Key Research and Development Program of China (2016YFB1001000), Key Research Program of Frontier Sciences, CAS (ZDBS-LY-JSC032), Shandong Provincial Key Research and Development Program (2019JZZY010119), and CAS-AIR.

{\small
	
	\bibliographystyle{plain}
%	\bibliography{egbib}
}

\end{document}